# PORCA: Modeling and Planning for Autonomous Driving among Many Pedestrians

Yuanfu Luo[1]  Panpan Cai[1]  Aniket Bera[2]  David Hsu[1]  Wee Sun Lee[1]  Dinesh Manocha[3]

*Abstract*—This paper presents a planning system for autonomous driving among many pedestrians. A key ingredient of our approach is PORCA, a pedestrian motion prediction model that accounts for both a pedestrian's global navigation intention and local interactions with the vehicle and other pedestrians. Unfortunately, the autonomous vehicle does not know the pedestrians' intentions a priori and requires a planning algorithm that hedges against the uncertainty in pedestrian intentions. Our planning system combines a POMDP algorithm with the pedestrian motion model and runs in real time. Experiments show that it enables a robot scooter to drive safely, efficiently, and smoothly in a crowd with a density of nearly one person per square meter.

*Index Terms*—Autonomous vehicle navigation; motion and path planning; planning under uncertainty; crowd motion models

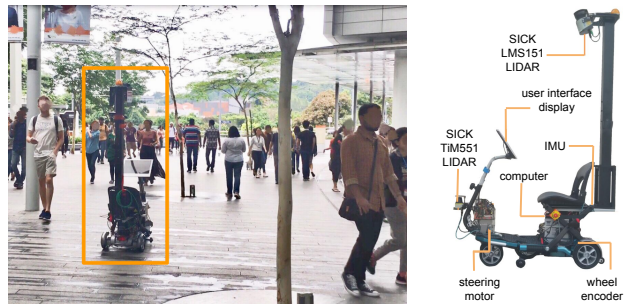

Fig. 1: Our robot scooter among many pedestrians on a campus plaza at the National University of Singapore.

## I. Introduction

DRIVING safely, efficiently, and smoothly among many pedestrians is a crucial capability of autonomous vehicles that operate in densely populated public places, such as airport terminals, shopping malls, hospital complexes etc.. It is challenging, because of the uncertainty of pedestrian motions, constantly changing environments, sensor noise, and imperfect robot control. See Fig. 1 for an example.

To drive successfully among pedestrians, the autonomous vehicle must understand pedestrian behaviors. Pedestrian movements generally depend on *intentions* and *interactions*. Intentions represent global navigation goals, e.g., final destinations. However, intentions alone do not fully determine pedestrian movements. Interactions with obstacles, vehicles, and other pedestrians in the local environment also significantly affect pedestrian movements. A pedestrian motion model must account for both intentions and interactions in order to capture pedestrian motions accurately at both global and local scales.

Integrating an accurate pedestrian motion model with a planning algorithm enables autonomous driving among pedestrians. Unfortunately, pedestrian intentions and interactions often contain significant uncertainty. Intentions of pedestrians are not directly observable and must be inferred. Their interactions vary widely across individuals, because of differences in age, gender, risk propensity, etc.. Handling these uncertainties is thus key to robust planning for driving among pedestrians.

Our planning system incorporates pedestrian intentions, interactions, and their uncertainties in a principled manner. We first develop a pedestrian motion model conditioned on both intentions and interactions. We then embed the model within a Partially Observable Markov Decision Process (POMDP) to plan for optimal vehicle control under uncertainty.

Our pedestrian motion model extends *Optimal Reciprocal Collision Avoidance* (ORCA) [1], which is a general motion model for multi-agent reciprocal collision avoidance. ORCA provides sufficient conditions for collision-free motion by assuming that each interacting agent takes half of the responsibility of avoiding pairwise collisions. Our extension, *Pedestrian ORCA* (PORCA), addresses two main limitations of ORCA. First, ORCA suffers from the *freezing pedestrian* problem, when applied to pedestrian motion modeling. Under the ORCA model, agents try to move along the direction of their preferred velocity as much as possible. When two pedestrians or a pedestrian and a vehicle encounter head-on, both insist on their direction of movement, instead of coordinating to take a small detour. To avoid collision, both must slow down, sometimes, to a complete stop, resulting in the freezing pedestrian problem. PORCA augments the agent's objective function in order to encourage it to move and thus explore new directions in order to escape from the "freezing state". Second, the original ORCA model assumes a homogeneous set of agents. For pedestrian-vehicle interaction, the vehicle is more restricted in local movement than the pedestrian, because of the non-holonomic constraints. To address this asymmetry, PORCA assigns greater responsibility to the pedestrian for

We thank Malika Meghjani, You Hong Eng, and Wei Kang Leong from the Singapore-MIT Alliance for Research & Technology (SMART) for their help with the robot scooter system. This research is supported in part by SMART IRG grant R-252-000-655-592, Singapore MoE grant MOE2016-T2-2-068, US ARO grant W911NF16-1-0085, and Intel Corporation.

[1]The authors are with the School of Computing, National University of Singapore. {yuanfu,caipp,dyhsu,leews}@comp.nus.edu.sg
[2]The author is with the Department of Computer Science, University of North Carolina at Chapel Hill, USA. ab@cs.umd.edu
[3]The author is with the Department of Computer Science and Electrical & Computer Engineering University of Maryland at College Park, USA. dm@cs.umd.edu

collision avoidance, when the pedestrian and the vehicle are close to each other.

We incorporate PORCA into a probabilistic model of system dynamics. Our POMDP model encodes pedestrian intentions as hidden variables and applies PORCA in the state-transition function to predict pedestrian motions conditioned on their intentions and interactions. By solving the POMDP, our system performs intention- and interaction-aware planning for the autonomous vehicle. We use a parallel version of the DESPOT algorithm [2] to solve the POMDP model efficiently. The algorithm maintains a *belief*, i.e., probability distribution, over possible intentions of each pedestrian. At each time step, it performs a lookahead search in a belief tree reachable under future actions and observations, to plan for optimal vehicle control.

Experiments show that PORCA predicts pedestrian motions more accurately than prior models. By integrating it with POMDP planning, our autonomous vehicle successfully avoids collisions with pedestrians and reaches the goal more efficiently and smoothly.

## II. RELATED WORK

### A. Planning for Navigation among Pedestrians

Considerable research has been conducted on navigation among pedestrians. Many previous planning algorithms ignore pedestrians' intentions and interactions when predicting their motions. For example, the approaches in [3], [4] treat pedestrians as static obstacles and handle pedestrian dynamics through online replanning. Other approaches assume simple independent motions for pedestrians, e.g., constant velocity [5]. Another group of planning algorithms consider pedestrians' intentions, but do not explicitly model their interactions. For example, Foka and Trahanias [6] have integrated navigation goals and short-term motions into a dynamic costmap, assuming that pedestrians move independently and do not interact with each other. Thompson, Horiuchi and Kagami [7] have followed a very similar approach, using probability grids to encode pedestrians' independent motions. Both [8] and [9] have learned pedestrian motion patterns from data to predict trajectories, then applied A* to compute a path for the vehicle using those trajectories. These data-driven approaches, however, can hardly generalize to novel environments [7]. Bai et al. [10] have modeled the navigation problem as a POMDP to handle uncertain pedestrian intentions, but still used a simple straight-line motion model for them. Some planning algorithms, though few, have modeled both intentions and interactions of pedestrians. However, they often overlook the underlying uncertainty. The planning in [11] models pedestrian-vehicle interactions using interacting Gaussian processes. However, their method assumes each pedestrian has a fixed navigation goal, hence does not capture the uncertainty on pedestrian intentions. Kuderer et al. [12] performed navigation by optimizing trajectories in a joint state space of all pedestrians and the vehicle. It assumes that pedestrians' intentions are fixed in a planning cycle. The method is also highly computationally expensive.

Our work is related to the intention-aware planning in [10], but our POMDP models both intentions and interactions, and their underlying uncertainty. Instead of assuming straight-line motions of pedestrians as in [10], we integrate a sophisticated pedestrian motion model into the POMDP to better characterize pedestrians' short-term motions.

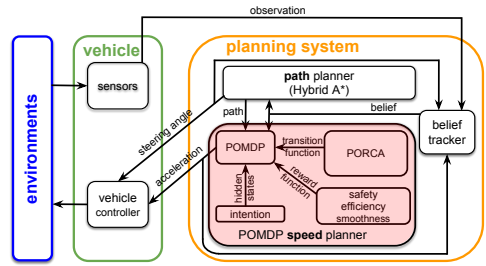

Fig. 2: System overview. The red box contains the key component of vehicle speed control.

### B. Pedestrian Motion Modeling

Pedestrian motion modeling has been studied extensively. There are three main categories of pedestrian motion models: social force based, data-driven and geometric approaches. Most existing methods do not explicitly model the interactions between pedestrians and non-holonomic vehicles. Social force models [13], [14], [15] assume that pedestrians are driven by virtual forces that measure the internal motivations of individuals for reaching the goal, avoiding obstacles, or performing certain actions, etc.. These approaches perform well in simulating crowds, but often predict the movements of individual pedestrian poorly [12]. Data-driven approaches [16], [17] learn pedestrian dynamics from past trajectories. However, the training data required is often hard to obtain. Besides, the learned models may not generalize well to novel scenarios. Geometric approaches compute collision-free paths for multiple agents via optimization in the feasible geometric space. This category includes the well-known Velocity Obstacle (VO) based [18] and the Reciprocal Velocity Obstacle (RVO) based algorithms [19], [1], [20].

Some previous work has tried to handle non-holonomic motion of vehicles. The method in [21] uses a trajectory tracking controller to generate non-holonomic vehicle motion after applying ORCA. However, inside ORCA, the vehicle is still considered as a holonomic agent. Models in [22], [23] handle pedestrian-vehicle interactions specifically at crosswalks but are not applicable for other general scenes.

Our pedestrian motion model is developed upon ORCA [1], a model based on optimal reciprocal collision avoidance. We improve ORCA by taking the non-holonomic nature of vehicles into account and enhancing its objective function to simulate more natural interactions.

## III. OVERVIEW

Fig. 2 shows an overview of our driving system. It consists of a belief tracker, a *path* planner, and a *speed* planner. The belief tracker maintains a belief over pedestrian intentions and constantly updates it to integrate new observations. The path planner uses hybrid A* [24] to plan a non-holonomic driving path, and extracts vehicle steering commands from the path. The speed planner solves an intention POMDP model to control the vehicle speed along the planned path. The planning system re-plans both the steering and the speed at 3 HZ.

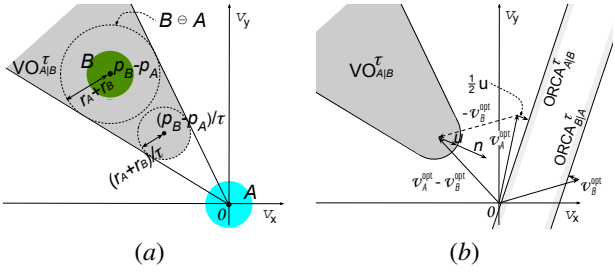

Fig. 3: (a) $\text{VO}_{A|B}^\tau$ (gray), the velocity obstacle of agent $A$ (blue) induced by agent $B$ (green) for time $\tau$, is a truncated cone with its apex at the origin (in velocity space) and its legs tangent to the disc $B \ominus A$. If the relative velocity of $A$ with respective to $B$ is in $\text{VO}_{A|B}^\tau$, $A$ and $B$ will collide with each other before time $\tau$. (b) $\text{ORCA}_{A|B}^\tau$ is a half-plane divided by the line that is perpendicular to the vector $u$ through the point $v_A^{\text{opt}} + \frac{1}{2}u$, where $u$ is the vector from $v_A^{\text{opt}} - v_B^{\text{opt}}$ to the closest point on the boundary of $\text{VO}_{A|B}^\tau$.

This work focuses on building the speed planner to achieve intention- and interaction-aware autonomous driving among pedestrians; we refer readers to [10] for more details on the path planner and the belief tracker. The red box in Fig. 2 presents a detailed view of our speed planner. We constructed a POMDP model encoding pedestrian intentions as hidden states. At each time step, the speed planner performs a look-ahead search in a belief tree, using a pedestrian motion model (PORCA) to predict pedestrian behaviors. The vehicle then executes the first action in the plan, which is optimized for safe, efficient and smooth driving among many pedestrians.

We will first introduce our pedestrian motion model in Section IV, and then introduce our POMDP speed planner that integrates this pedestrian motion model in Section V.

## IV. PEDESTRIAN MOTION PREDICTION

The pedestrian motion model takes as input the intentions of pedestrians and the vehicle, and simulates their interactions to predict pedestrian motions. We improve ORCA [1], a pedestrian simulation algorithm based on reciprocal collision avoidance, to model more natural pedestrian-vehicle and pedestrian-pedestrian interactions.

For completeness, we will first introduce ORCA and its limitations, and then present our model that addresses these limitations.

### A. Optimal Reciprocal Collision Avoidance

For a given agent, e.g., a pedestrian or a vehicle, ORCA generates half-planes of velocities that allow it to avoid collision with other agents. It then selects the optimal velocity for the given agent from the intersection of the half-planes using linear programming. ORCA computes the half-planes based on Velocity Obstacle (VO) [18].

*1) Velocity Obstacles:* For two agents $A$ and $B$, the velocity obstacle $\text{VO}_{A|B}^\tau$ is defined as the set of all *relative* velocities of $A$ with respect to $B$ that will result in a collision between $A$ and $B$ before time $\tau$. Formally, $\text{VO}_{A|B}^\tau$, the *velocity obstacle* of agent $A$ induced by $B$ with time window $\tau$ is defined as:

$$\text{VO}_{A|B}^\tau = \{v | \delta(p_A, v, \tau) \cap (B \ominus A) \neq \emptyset\}. \quad (1)$$

The Minkowski difference $B \ominus A$ inflates the geometry of B by that of A, so that $A$ can be treated as a single point. For simplicity, ORCA represents all the agents as disc shapes. The Minkowski difference $B \ominus A$ then becomes a disk of radius $r_A + r_B$, where $r_A$ and $r_B$ are the radius of $A$ and $B$, respectively. $\delta(p_A, v, \tau)$ is the straight-line relative trajectory traveled by $A$ with respect to $B$ during time $(0, \tau)$, by starting from its position $p_A$, and taking the relative velocity $v$. Suppose $A$ and $B$ are moving with velocity $v_A$ and $v_B$, respectively. If $v_A - v_B \notin \text{VO}_{A|B}^\tau$, $A$ and $B$ are guaranteed to be collision-free for at least $\tau$ time. See Fig. 3a for the geometric interpretation of velocity obstacle.

*2) Collision-Avoiding Velocity Set:* Suppose $B$ selects its velocity $v_B$ from a set $V_B$. To avoid the collision for at least $\tau$ time, $A$ needs to choose its velocity $v_A$ so that $v_A \notin \text{VO}_{A|B}^\tau \oplus V_B$, where $\oplus$ denotes the Minkowski sum. This leads to the definition of *collision-avoiding velocity set* of $A$ with respect to $B$:

$$\text{CA}_{A|B}^\tau(V_B) = \{v | v \notin \text{VO}_{A|B}^\tau \oplus V_B\}. \quad (2)$$

A pair of velocity sets $V_A$ and $V_B$ is called *reciprocal collision-avoiding* if $V_A \subseteq \text{CA}_{A|B}^\tau(V_B)$ and $V_B \subseteq \text{CA}_{B|A}^\tau(V_A)$; it is further called *reciprocal maximal* if $V_A = \text{CA}_{A|B}^\tau(V_B)$ and $V_B = \text{CA}_{B|A}^\tau(V_A)$.

*3) Optimal Reciprocal Collision-Avoiding Velocity Set:* ORCA attempts to find the pair of reciprocal maximal velocity sets for each pair of agents $A$ and $B$, with the guidance of the *optimization velocities* $v_A^{\text{opt}}$ for $A$ and $v_B^{\text{opt}}$ for $B$ (Fig. 3b), such that the pair of velocity sets maximize the amount of permitted velocities close to $v_A^{\text{opt}}$ and $v_B^{\text{opt}}$. The optimization velocities $v_A^{\text{opt}}$ and $v_B^{\text{opt}}$ can be either the preferred velocities of $A$ and $B$ which correspond to their intentions, or simply set as their current velocities. The target pair of velocity sets, denoted as $\text{ORCA}_{A|B}^\tau$ for $A$ and $\text{ORCA}_{B|A}^\tau$ for $B$, can be constructed as follows [1].

Suppose that $v_A^{\text{opt}} - v_B^{\text{opt}} \in \text{VO}_{A|B}^\tau$, i.e., $A$ and $B$ will collide with each other before time $\tau$ by taking these velocities. To achieve collision avoidance with the least effort, ORCA finds a relative velocity from the boundary of $\text{VO}_{A|B}^\tau$ that is closest to $v_A^{\text{opt}} - v_B^{\text{opt}}$. Let $u$ be the vector from $v_A^{\text{opt}} - v_B^{\text{opt}}$ to this point:

$$u = (\underset{v \in \partial \text{VO}_{A|B}^\tau}{\arg\min} ||v - (v_A^{\text{opt}} - v_B^{\text{opt}})||) - (v_A^{\text{opt}} - v_B^{\text{opt}}). \quad (3)$$

Then $u$ is the smallest change on the relative velocity to avoid the collision within $\tau$ time. ORCA lets each agent take half of the responsibilities for collision avoidance, i.e., adapt its velocity by (at least) $\frac{1}{2}u$. It constructs $\text{ORCA}_{A|B}^\tau$, the optimal

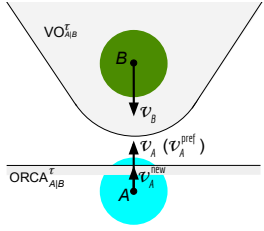

Fig. 4: An example showing the freezing pedestrian problem. Two pedestrians ($A$ and $B$) are walking towards each other, with velocities $v_A$ and $v_B$, respectively. Suppose $v_A^{\text{pref}} = v_A$. The new velocity $v_A^{\text{new}}$ computed by ORCA is as shown. Its magnitude is much smaller than that of $v_A^{\text{pref}}$, which means $A$ slows down a lot. This behavior is different from that of human. Human usually maintains a similar speed (magnitude) and makes a detour to avoid collisions instead of slowing down a lot.

collision-avoiding velocity set of agent $A$ with respect to $B$, as:

$$\text{ORCA}_{A|B}^\tau = \{v | (v - (v_A^{\text{opt}} + \frac{1}{2}u)) \cdot n \geq 0\}, \quad (4)$$

where $n$ is the outward normal at point $(v_A^{\text{opt}} - v_B^{\text{opt}}) + u$ on the boundary of $\text{VO}_{A|B}^\tau$. $\text{ORCA}_{B|A}^\tau$ is constructed symmetrically (Fig. 3b).

Geometrically, $\text{ORCA}_{A|B}^\tau$ and $\text{ORCA}_{B|A}^\tau$ are half-planes of velocities in the velocity space (Fig. 3b).

*4) Computing the New Velocity:* For an agent $A$, ORCA computes for each other agent $B$ the optimal reciprocal collision-avoiding velocity set $\text{ORCA}_{A|B}^\tau$. The permitted velocity set for $A$, denoted as $\text{ORCA}_A^\tau$, is the intersection of the half-planes $\text{ORCA}_{A|B}^\tau$ induced by all $B$'s:

$$\text{ORCA}_A^\tau = \bigcap_{B \neq A} \text{ORCA}_{A|B}^\tau. \quad (5)$$

The agent $A$ then selects a new velocity $v_A^{\text{new}}$ that is closest to its preferred velocity $v_A^{\text{pref}}$ from $\text{ORCA}_A^\tau$, i.e.,

$$v_A^{\text{new}} = \arg\min_{v \in \text{ORCA}_A^\tau} ||v - v_A^{\text{pref}}||. \quad (6)$$

The computation of $v_A^{\text{new}}$ can be efficiently done using linear programming [1].

### B. Limitations of ORCA

The original ORCA has two limitations: the *freezing pedestrian* problem and the *violation of non-holonomic constraints* problem.

*1) Freezing Pedestrians:* The objective function (6) for ORCA attempts to minimize the distance from the new collision-free velocity to the pedestrian's preferred velocity that points to his/her goal position. This sometimes causes the "freezing pedestrian" problem in pedestrian simulations. Pedestrians insist on walking in their goal directions, causing them to walk very slowly or even stay stationary when obstructed along their way. Pedestrians in reality, however, would like to maintain their speed (the magnitude of the velocity) and make necessary detours to bypass the obstacles in such scenarios. See Fig. 4 for an example of this problem.

*2) Violation of Non-holonomic Constraints:* ORCA assumes the motions of agents are holonomic and agents can flexibly change their moving directions to avoid collisions. It is true for pedestrians. Most vehicles in reality, however, are under the non-holonomic kinematic constraints and cannot freely change their directions. With holonomic assumptions for vehicles, pedestrians in ORCA will have wrong anticipations on the vehicle's motions during reciprocal collision avoidance. This often leads to unnatural simulations of pedestrians. For example, in ORCA, when a pedestrian is close to a vehicle, he/she still thinks that the vehicle can flexibly avoid him/her by moving side-wise, thus will not change his/her walking direction much. In reality, a pedestrian in this dangerous situation would try his/her best to avoid collisions.

### C. PORCA

In this section, we present our extension, PORCA, which addresses the aforementioned problems.

*1) Objective Function with Patience:* The freezing pedestrian problem could be addressed by several approaches, such as performing multiple-step lookahead in ORCA, or applying a global planner to replan preferred velocities for pedestrians. However, these approaches would significantly increase the computation time of ORCA, and are thus not suitable for our application, because ORCA will be used heavily during each planning cycle. In this section, we propose a solution to the freezing pedestrian problem that has rather low computational overhead.

Concretely, we design a new objective function for ORCA as follows

$$v_A^{\text{new}} = \arg\min_{v \in \text{ORCA}_A^\tau} \left\{ ||v - v^{\text{pref}}||^2 + \frac{1}{\varrho_A} \left| ||v||^2 - ||v^{\text{pref}}||^2 \right| \right\}, \quad (7)$$

where the variable $\varrho_A \in (0, 1]$ measures the *patience* of a pedestrian $A$. We observed that, pedestrians usually prefer to maintain their current speed. If they intend to move, they often get more and more impatient when staying put or walking very slowly. Therefore, we introduce the second term in (7) to penalize unintended slowing down of pedestrians. We use the pedestrian's *patience* level $\varrho$ to control the strength of this penalization. Intuitively, if a pedestrian walks at a low speed, it gets more and more impatient ($\varrho$ decreases) over time. The weight $\frac{1}{\varrho}$ for the second term in (7) then increases and encourages the pedestrian to explore alternative directions that enable him/her to move faster.

We set the patience for each pedestrian intending to move as follows. It is set to 1 initially, which indicates a normal level of patience, and starts to decrease exponentially with the time if the pedestrian's new speed is less than some threshold $\varsigma$ (set to $0.2||v^{\text{pref}}||$ in our experiments). When the patience is smaller than $\varrho_{\min}$ ($\varrho_{\min} = 0.1$ in our experiments), we set it to $\varrho_{\min}$; it is reset to 1 once the new speed exceeds $\varsigma$. Empirically, we suggest to set $\varsigma \in (0, 0.3||v^{\text{pref}}||)$ and $\varrho_{\min} \in (0.05, 0.15)$. For pedestrians intending to stay put, we fix their patiences to be 1.

One may concern that the objective function (7) is no longer convex when $\varrho < 1$ and that increases the computational cost of PORCA. From our observation, fortunately, pedestrian speeds are smaller than $\varsigma$ only occasionally, meaning that $\varrho$ is 1 for most of the time. Hence we can still solve PORCA very efficiently.

*2) ORCA with Changing Responsibilities:* To handle the non-holonomic motion of the vehicle and produce more natural pedestrian-vehicle interactions, we let pedestrians take more responsibilities for reciprocal collision avoidance when they are very close to a vehicle. The vehicle takes less responsibilities accordingly. For the vehicle, this mechanism selects velocities that are closer to its current velocity, and thus better complies with the non-holonomic constraint. For pedestrians, it models their urgency to avoid collisions with the non-holonomic vehicle.

The responsibility $r \in (0, 1)$ of an agent is defined as the ratio of the velocity $u$ (defined in (3)) this agent takes to avoid collision with others. If an agent $A$ takes $r$ responsibility to avoid the collision with $B$, its $\text{ORCA}_{A|B}^{\tau}$ will be computed as,

$$\text{ORCA}_{A|B}^{\tau} = \{v | (v - (v_A^{\text{opt}} + ru)) \cdot n \geq 0\},$$

$B$ will have $\text{ORCA}_{B|A}^{\tau}$ with responsibility $1 - r$ accordingly. Initially, the pedestrian and the vehicle each take half of the responsibilities. When their distance is within some threshold $d$, the responsibility of the pedestrian increases linearly as he/she approaches the vehicle. The responsibility reaches a maximum value $R$ when the distance decreases to zero. We set $d = 1.5$ and $R = 0.95$ in our experiments. We found empirically that the linear function works well.

## V. INTENTION-AWARE AND INTERACTION-AWARE AUTONOMOUS DRIVING

Our planning system uses a two-level hierarchical approach [10] for autonomous driving to reduce the computational cost. At the high level, hybrid A* [24] is used to plan a path; at the low level, a POMDP model is built to control the vehicle *speed* along the planned path. This section focuses on the low-level POMDP speed planner.

### A. POMDP Preliminaries

Formally, a POMDP is defined as a tuple $(S, A, Z, T, O, R, b_0)$, where $S$, $A$ and $Z$ represent the state space, the action space and the observation space, respectively. The transition function $T(s, a, s') = p(s'|s, a)$ models the imperfect robot control and environment dynamics. It defines the probability of transiting to a state $s'$ from a state $s$, after the robot executes an action $a$. The observation function $O(s, a, z) = p(z|a, s)$ characterizes the robot's sensing noises. It defines the probability of receiving the observation $z$ after the robot executes $a$ and reaches $s$. The reward function $R(s, a)$ defines a real-valued immediate reward for executing $a$ at $s$. Due to imperfect sensing, the robot does not know the exact state of the world. Instead, it maintains a *belief*, which is a probability distribution over $S$, and reasons over the space of beliefs. At each time step, the belief is updated via the Bayes' rule:

$$b_t(s') = \eta O(s', a_t, z_t) \sum_{s \in S} T(s, a_t, s') b_{t-1}(s), \quad (8)$$

where $\eta$ is a normalizing constant.

POMDP planning aims to find a *policy* $\pi$, a mapping from a belief $b$ to an action $a$, that maximizes the expected total discounted rewards:

$$V_\pi(b) = \mathbb{E}\left( \sum_{t=0}^{\infty} \gamma^t R(s_t, \pi(b_t)) \,\Big|\, b_0 = b \right), \quad (9)$$

where $s_t$ is the state at time $t$, $\pi(b_t)$ is the action that the policy $\pi$ chooses at time $t$, and $\gamma \in (0, 1)$ is a discount factor that places preferences for immediate rewards over future ones. The expectation is taken over the sequence of uncertain state transitions and observations over time.

POMDP planning is usually performed as a lookahead search in a *belief tree*. Each node of the belief tree corresponds to a belief. At each node, the tree branches on all actions and observations. Beliefs in child nodes are computed using the Bayes' rule (8). The output of the search is an optimal policy conditioned on the initial belief.

### B. Intention-Aware POMDP for Autonomous Driving

*1) State Modeling:* A state in our model consists of two parts: the vehicle state and the pedestrian state. The vehicle state consists of its 2D position $(x, y)$, heading direction $\theta$, and instantaneous speed $v$. The state of a pedestrian consists of its position $(x, y)$, speed $v$ and intention $g$. The intention of a pedestrian is represented as his/her goal location in the environment. Pedestrians' intentions are not observable to the car, and thus are treated as hidden variables in the state and must be inferred from the history of interactions.

*2) Action Modeling:* Our model uses discrete actions: at each time step, the planning can choose to ACCELERATE, DECELERATE, or MAINTAIN the current vehicle speed, in order to avoid collisions and navigate efficiently and smoothly towards the goal.

*3) Observation Modeling:* An observation in our model consists of the vehicle position, its speed, and the positions of all pedestrians. These observations are relatively accurate and their noises do not pose significant effect on decision making. Therefore, we assume them to be fully observable, and focus more on modeling the uncertainty in pedestrians' hidden intentions.

*4) Transition (Intention and Interaction) Modeling:* We decompose the transition function into two parts: vehicle transition and pedestrian transitions.

In each transition step, the vehicle takes a discrete action, and drives for a fixed time duration $\Delta t$ along the path planned by hybrid A*. We add a small Gaussian noise to the transition to model imperfect vehicle control. The resulting vehicle transition function, denoted as $p(x_{t+1}, y_{t+1}, v_{t+1} | x_t, y_t, v_t, a)$, satisfies the non-holonomic constraints.

Pedestrians' transitions are calculated using PORCA. Given a set of intentions, $(g^1, g^2, ..., g^n)$, of $n$ surrounding pedestrians, we compute for each pedestrian a preferred velocity pointing to his/her goal and with a speed assumed to be roughly the human-walking speed, $1.2 m/s$. These preferred velocities, together with positions of pedestrians and the vehicle, are input to PORCA to compute new velocities and predict next-step positions of pedestrians. To model stationary pedestrians, a "stop intention" (with preferred velocity 0) is also included. Gaussian noises are added on those predicted positions to model uncertainty in pedestrian-vehicle and pedestrian-pedestrian interactions. This pedestrian transition model is denoted as:

$$p(x_{t+1}^1, y_{t+1}^1, ..., x_{t+1}^n, y_{t+1}^n | x_t^1, y_t^1, ..., x_t^n, y_t^n, \\ g^1, ..., g^n, x_t, y_t, v_t) \quad (10)$$

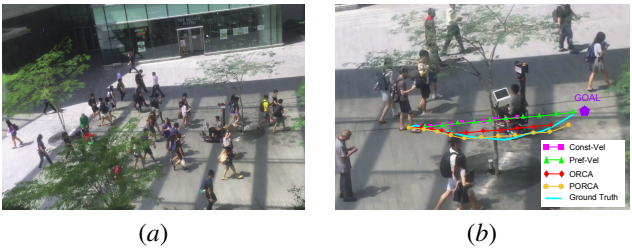

Fig. 5: Comparing pedestrian motion models. (*a*) The environment for data collection. (*b*) Predicted trajectories using four different motion models, compared with the ground truth.

TABLE I: The success rates of four motion models for pedestrian trajectory prediction.

| Const-Vel | Pref-Vel | ORCA | PORCA |
|---|---|---|---|
| 0.521 | 0.674 | 0.760 | 0.804 |

This pedestrian transition function (10) is conditioned on a set of given pedestrian intentions. The uncertainty in pedestrian intentions is systematically handled by the belief tree search and belief update in our POMDP planning.

*5) Reward Modeling:* The reward function encourages the vehicle to drive safely, efficiently, and smoothly. For safety, we give a huge penalty $R_{\text{col}} = -1000 \times (v^2 + 0.5)$, varying with the driving speed $v$, to the vehicle if it collides with any pedestrian. For efficiency, we assign a reward $R_{\text{goal}} = 0$ to the vehicle when it reaches the goal, and assign a penalty $R_{\text{speed}} = \frac{v - v_{\max}}{v_{\max}}$ to encourage the vehicle to choose a speed $v$ closer to its maximum speed $v_{\max}$, when it is safe to do so. For smoothness of the drive, we add a small penalty $R_{\text{acc}} = -0.1$ for the actions ACCELERATE and DECELERATE, to penalize the excessive speed changes.

### C. Solving the Intention-Aware POMDP

We use a parallel version of DESPOT [2] to efficiently solve the intention POMDP. The algorithm performs online POMDP planning through parallel belief tree search and parallel Monte Carlo simulations at leaf nodes of the belief tree. Within a simulation step in the planning, we further parallelize the transitions of individual pedestrians. Benefiting from the computational efficiency of the parallelized DESPOT, our planning system is able to re-plan the vehicle speed at 3Hz.

## VI. EXPERIMENTS

In the experiments, we first evaluate the prediction accuracy of our pedestrian motion model using real-world data. Then, we illustrate the performance of our planning system on three challenging scenarios in simulation. Finally, we show that our planning system can successfully drive a real robot vehicle among pedestrians.

### A. Pedestrian Motion Prediction

To test the prediction accuracy, we extracted 2D trajectories from real-world pedestrian videos shot in a campus plaza (Fig. 5*a*), and compared the predicted trajectories with them. We manually labeled 46 trajectories, each consisting of a discretized sequence of ground-truth positions for a particular pedestrian. Each position in a trajectory corresponds to one time frame. The duration between two adjacent frames is $0.33$ second. We then applied our model to predict the trajectories of the real pedestrians for a duration of $3$ seconds. We compute the distance between the predicted position at each frame of a trajectory and the ground truth at the same frame. The prediction is counted as a success if the average distance over all frames is less than $d$ meters. We set $d = 0.4$ to have the average one-second error smaller than $1.2$ meters, roughly the stride length of a pedestrian.

We compared our success rate with those of other motion models: constant velocity (Const-Vel), preferred velocity (Pref-Vel), and the original ORCA. Const-Vel assumes each pedestrian keeps his/her current velocity; Pref-Vel assumes each pedestrian walks towards his/her goal at a constant speed. Pref-Vel, ORCA and PORCA require goal information to predict pedestrian motions. Therefore, we extracted the ground truth goals from the trajectories and used them in all the models for fair comparisons. Results in TABLE I shows that our model achieved higher success rate than other models.

For a detailed view of the performance, we selected an example scene and visualized the predictions and the ground truth in Fig. 5*b*. In this scene, a pedestrian walking towards GOAL is blocked by a moving vehicle. Both Const-Vel and Pref-Vel predict that the pedestrian will walk in a straight line, ignoring that such trajectories lead to collisions. Both ORCA and PORCA predict that the pedestrian will detour. However, only our model successfully predicts that the pedestrian will maintain his speed during the detour.

### B. Autonomous Driving in Simulation

We analyze the performance of our planning algorithm (POMDP-PORCA) by comparing it with two types of baselines. The first type includes constant speed (Const-Speed), reactive controller (Reactive-Controller), and the POMDP speed planner with Pref-Vel model (POMDP-Pref-Vel). They only control the speed, and rely on hybrid A* to generate steering commands. The other type includes dynamic hybrid A* (Dynamic-Hybrid-A*), which controls both the steering and the speed. All the baselines do not model at least one of the following key aspects: intentions, interactions, and uncertainty. By comparing our algorithm with these baselines, we analyzed the benefit of modeling these aspects. Now we describe these baselines in detail.

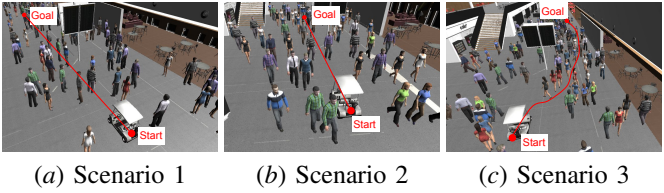

(a) Scenario 1    (b) Scenario 2    (c) Scenario 3

Fig. 6: Three handcrafted scenarios for evaluating autonomous driving in a crowd. Scenario 1: all pedestrians stay stationary; Scenario 2: all pedestrians walking towards the vehicle; Scenario 3: 150 pedestrians walking towards seven different goals. In Scenarios 1 and 2, the vehicle is required to drive along a *straight line* for 16 meters. In Scenario 3, the vehicle path is planed in real time.

TABLE II: Performance comparison. For each scenario, we run for 300 times. The average collision rate, average travel time, and average number of accelerating and decelerating actions are calculated using only the successful trials.

| Scenario | Algorithm | Collision Rate | Travel Time (s) | Accel/ Decel Number | Success Rate |
|---|---|---|---|---|---|
| 1 | Const-Speed | 0.177 | 24.19 | 0.0 | 1.0 |
|   | Reactive-Controller | - | - | - | 0.0 |
|   | POMDP-Pref-Vel | 0.0 | 27.27 | 46.6 | 0.97 |
|   | POMDP-PORCA | 0.0 | 24.19 | 45.1 | 1.0 |
| 2 | Const-Speed | 0.063 | 34.63 | 0.0 | 1.0 |
|   | Reactive-Controller | 0.0 | 75.72 | 128.0 | 1.0 |
|   | POMDP-Pref-Vel | 0.0 | 62.57 | 141.4 | 0.977 |
|   | POMDP-PORCA | 0.0 | 34.63 | 104.8 | 1.0 |
| 3 | Const-Speed | 0.033 | 71.38 | 0.0 | 1.0 |
|   | Dynamic-Hybrid-A* | 0.023 | 50.46 | 36.1 | 1.0 |
|   | Reactive-Controller | 0.0 | 183.11 | 128.8 | 0.817 |
|   | POMDP-Pref-Vel | 0.0 | 105.20 | 190.7 | 0.91 |
|   | POMDP-PORCA | 0.0 | 74.36 | 157.6 | 0.963 |

Const-Speed drives the vehicle at a constant speed. We set the speed as the average speed of POMDP-PORCA, to compare the safety when it achieves the same efficiency as our algorithm. Reactive-Controller avoids collisions with pedestrians without modeling intentions and interactions. It compares $D$, the distance to the nearest pedestrian, with two distance thresholds $D_{\text{far}}$ and $D_{\text{near}}$, and then chooses DECELERATE if $D < D_{\text{near}}$, ACCELERATE if $D > D_{\text{far}}$, or MAINTAIN if $D_{\text{near}} < D < D_{\text{far}}$. POMDP-Pref-Vel is similar to our algorithm, except that it uses Pref-Vel instead of PORCA as the pedestrian motion model, considering only intentions. Dynamic-Hybrid-A* is hybrid A* augmented with an acceleration dimension in the search space and with predicted pedestrian positions in the collision checking module. Pedestrian motions are predicted using a constant velocity model. Dynamic-Hybrid-A* also does not explicitly model intentions and interactions.

The criteria for comparison include safety, efficiency and smoothness. We measure the safety by the collision rate, the efficiency by the success rate and the travel time, and the smoothness by the number of accelerations and decelerations. We ran 300 trials for each scenario and computed the success rate; a trial is considered as successful if the vehicle reaches its goal within 6 minutes. We computed the average collision rate, travel time and accel/decel number *using only the successful trials*.

We tested the algorithms in three simulated scenarios (Fig. 6). The first two (Fig. 6a and b) are scenarios where modeling intentions and interactions are especially important. Scenario 3 (Fig. 6c) represents a complex scene that is common in real life.

We built our simulator with the Unity game engine, and implemented a package of sensors, including 2D LIDARs, wheel encoders, etc.. The simulator uses PORCA to simulate pedestrian motions. The simulator communicates with our planning system using the Robot Operating System (ROS). Note that though this simulator is designed for testing our planning system, it is general and can be used in other autonomous driving applications involving pedestrians.

TABLE II shows the performance of the tested algorithms. Overall, our algorithm, POMDP-PORCA, guarantees safety, while Const-Speed and Dynamic-Hybrid-A* result in collisions, and it outperforms the other baselines that also achieve safety, Reactive-Controller and POMDP-Pref-Vel, in efficiency and smoothness.

Const-Speed drives the vehicle aggressively in all three scenarios; this leads to collisions when pedestrians fail to avoid the vehicle. Dynamic-Hybrid-A* also causes collisions. Without modeling intentions and interactions, it fails to accurately predict pedestrian motions. Moreover, since no uncertainty is considered, it drives the vehicle aggressively. Reactive-Controller drives the vehicle over-conservatively in all scenarios because it does not model the interactions of reciprocal collision avoidance. For example, in Scenario 1, Reactive-Controller keeps the vehicle waiting for stationary pedestrians in front, in order to guarantee safety. Hence the vehicle never reaches its goal (0 success rate). With our algorithm, the vehicle knows that pedestrians will cooperate with it to avoid collisions. Therefore, instead of waiting, the vehicle slowly moves forward to convey its intention. Pedestrians hence give way to the vehicle. POMDP-Pref-Vel can achieve relatively higher success rates than Reactive-Controller, but still requires much more travel time and drives less smoothly than our algorithm.

### C. Autonomous Driving with A Robot Scooter

We further tested the performance of our planning system on a real robot vehicle (Fig. 1).

The sensor package of our robot vehicle includes two LIDARs, an Inertial Measurement Unit (IMU), and wheel encoders. The top-mounted SICK LMS151 LIDAR and the bottom-mounted SICK TiM551 LIDAR are used for pedestrian detection and localization, respectively. Our planning system runs on an Ethernet-connected laptop with an Intel Core i7-4770R CPU running at 3.90 GHz, a GeForce GTX 1050M GPU, and 16 GB main memory. The maximum speed for the vehicle is set to $1m/s$ for safety. The planning system is implemented on ROS. Our vehicle detects pedestrians from laser points using $K$-means clustering, and tracks pedestrians between two adjacent time frames by comparing the difference of their corresponding laser clusters. It localizes itself in a given map using adaptive Monte Carlo localization [25], which integrates information from the LIDAR, the IMU, and the wheel encoders.

We tested our autonomous driving system on a campus plaza (Fig. 1) for multiple times. Overall, our planning system performs well. The vehicle achieved its goal efficiently, smoothly and avoid the pedestrians successfully in all trials. See the video at http://motion.comp.nus.edu.sg/2018/06/23/autonomous-driving-in-a-crowd/ for more details.

## VII. Conclusion and Future Work

We developed an online planning system for autonomous driving in a crowd that considers both intentions and interactions of pedestrians. Our planning system combines a pedestrian motion model and a POMDP algorithm seamlessly to plan optimal vehicle actions under the uncertainty in pedestrian intentions and interactions. Our pedestrian motion model improves over previous models on the accuracy of predicting pedestrian interactions in the presence of the vehicle. By utilizing this prediction model, our planning system enables robot vehicles to drive safely, efficiently and smoothly among many pedestrians.

There are multiple directions we can work on in the future. First, we plan to better incorporate the non-holonomic constraints of vehicles into the computation of ORCA velocity sets, instead of using only changing responsibilities. Second, we can apply more sophisticated pedestrian motion models, such as GLMP [26], to improve the simulator. Finally, our POMDP model is actually a *POMDP-lite* [27], a subclass of POMDPs where the hidden state variables are constant or only change deterministically. Our POMDP model can be potentially solved more efficiently using POMDP-lite.